\documentclass[12pt]{article}
\usepackage{graphicx}%,float,url,graphics}
\usepackage{braket}
\usepackage{scalerel,amssymb}
\usepackage{amsmath}
\usepackage{comment}
\usepackage{placeins}
\usepackage[square,numbers,compress]{natbib}
\usepackage{authblk}

\begin{document}

\title{%Hopfield Neural Network method invariance with respect to conformable fractional derivative \\
%Applying the conformable fractional derivative in the Hopfield Neural Network method%: one may corrupting the conformable fractional derivative to get lesser learning time?
%The conformable fractional Hopfield Neural Network method
%The conformable fractional Hopfield Neural Network
The G\^ateaux-Hopfield Neural Network method
%\\
%Gateaux fractional Hopfield Neural Network
}

\author[1]{F. S. Carvalho\footnote{felipe.s.carvalho\_qui@hotmail.com}}
\author[1]{J. P. Braga\footnote{jjppbraga@gmail.com}}

\affil[1]{Departmento de Qu\'imica - ICEx, Universidade Federal de Minas Gerais, Belo Horizonte, Minas Gerais, Brazil.}

\maketitle
\clearpage

\begin{abstract}
%	{\sf 
%	Hopfield neural network were generalized within the Linear Extended Gateaux Derivative and referred as the  G\^ateaux Hopfiel neural network (GHNN).
%	 
%A new set of Hopfield differential equations in Hopfield neural network (HNN) method were established by means of the Linear Extended Gateaux Derivative (LEGD). Thus, this new method will be referred to as G\^ateaux Hopfiel neural network (GHNN). 
%A particular case, based on the Khalil conformable fractional derivative were discussed and applied to a first order Fredholm integral problem. The results 
%showed that the GHNN (for $\alpha>1$) is much faster, by a factor of 20, if compared to the integer order Hopfield equations . 
%Also, a limit to the learning time is observed by analysing the results for different values of $\alpha$. 
%The robustness and advantages of this new method is evident.}
%\\%
%
In the present work a new set of differential equations for the Hopfield Neural Network (HNN) method were established by means of the Linear Extended Gateaux Derivative (LEGD). This new approach will be referred to as G\^ateaux-Hopfiel Neural Network (GHNN). 
%A particular case for this definition was also discussed,
%choosen based on the Khalil conformable fractional derivative, which is obtained 
%imposing $\psi=t^{1-\alpha}$ and $\alpha\le1$ on LEGD. 
A first order Fredholm integral problem was used to test this new method and it was found to converge 22 times faster to the exact solutions for $\alpha>1$ if compared with the HNN integer order differential equations. Also a limit to the learning time is observed by analysing the results for different values of $\alpha$. The robustness and advantages of this new method will be pointed out.\\

\end{abstract}

\baselineskip=5.0ex

%\mathversion{bold}
%\clearpage
\section{Introduction}
%\subsection{Definindo e revendo problemas inversos}
Retrieving physical and chemical information from experimental data represents an important class of problem in science, known as inverse problems \cite{tik,jp1}. 
Due to the its inherent ill posed nature, inverse problems have to be regularized and solved by  special numerical techniques, such as the singular value decomposition \cite{leon}, the Tikhonov regularization \cite{tik2} or Hopfield  Neural Network (HNN) \cite{hop,hop2}. 
Examples of inverse problems solved by Hopfield Neural Network (HNN) can be, for example, the recovery of molecular force field from experimental frequencies \cite{jp2}, the probability density function from experimental positron annihilation lifetime spectra \cite{jp3} and applications on scattering theory \cite{bao,ammari}. Although the applications of HNN to inverse problem has proved to be very robust, considerable time can be saved if the concept of a more general derivative is introduced in the neural network framework. \\

%\subsection{Introduzir derivadas fracionarias, com enfase no Khalil}
Fractional derivative can be defined in a variety of ways. 
%Several definitions for fractional derivatives can be found in literature 
For a theoretical and historical review one may refer to references\cite{podlubny,miller1993,tarasov2013}. However, a large number of these definitions does not follow the usual properties of calculus, such as the derivative of constant not equal to zero or the properties of the chain rule. 
This will impose serious difficult to conciliate the definitions of fractional calculus with the analysis of physical and chemical problems, such as the learning time in the HNN. \\
%This will imply in 
%serious difficults to analyse physical and chemical problems. 

Among the several fractional derivative definitions, the Caputo fractional derivative has been largely used in the literature \cite{baleanu,murio,almeida}, although this definition 
%it has a serious shortcoming for it 
also does not satisfy the usual calculus chain rule. 
Even with this serious shortcoming
%Nevertheless, the 
Caputo fractional derivative has been applied to inverse problems in the literature \cite{zhang,kaslik,zhang2,boroomand}. 
%Fractional order derivative applied to inverse problems has been presented before in the literature \cite{zhang,kaslik,zhang2,boroomand}. These works use the Caputo fractional derivative to rewrite the HNN set of differential equations. Nevertheless this fractional order derivative has a serious issue. 
Dissipation of energy is not guaranteed in this framework.\\
%, since the usual chain rule is not obeyed.\\

One definition that satisfies the properties of calculus was proposed by Khalil {\it et. al.} and named as Conformable Fractional Derivative (CFD)\cite{khalil}. Since this definition is a Linear Extended G\^ateaux Derivative (LEGD) particular case\cite{zhao}, the Hopfield set of differential equations can be rigorous established in a more general sense. In this case dissipation of energy comes out in a natural way and the basic neural network equations are also simple modified. This more general result will be referred here as 
%is referred to as 
the G\^ateaux-Hopfield Neural Network (GHNN). \\

%On the other hand, Hopfield convergence on learning time can be rigorous established in the fractional derivative framework, if Khalil {\it et. al.} fractional derivative is used. 

A theoretical background is presented in the next section of the paper. This will introduce concepts usefull to prove relations along the work. Afterwards,the LEGD will be used to derive the Hopfield Neural Network equations, based on a previous integer order HNN\cite{jpinv} algorithm. A first order Fredholm integral equation will be used to test the new method. 
%
%The $\psi$ function in LEGD definition will be set as in CFD{\sf ... DEFINIU?}, $\psi=t^{1-\alpha}$, but with $\alpha \in \mathbb{R}$ instead of $\alpha<1$, {\it i. e.} it will allow greater freedom for $\alpha$ values. 
In the LEGD one must define a parameter $\psi$, which is the generalization of the derivation path in the directional derivative. It will be set as in CFD, $\psi=t^{1-\alpha}$, but with $\alpha \in \mathbb{R}$ instead of $\alpha<1$, allowing greater freedom for $\alpha$ values.
It was also possible to write a variation of the Euler method which will be used along the work. The present framework proves to be an improvement of the usual HNN method since the computational time is reduced significantly for large values of $\alpha$.

\section{Mathematical background}
The Linear Extended G\^ateaux derivative is defined as\cite{zhao} 

\begin{equation}\label{eq:legd2}
\begin{gathered}
df^{\text{leg}}(t;\psi) = \lim\limits_{\epsilon\rightarrow0} \frac{f(t+\epsilon\psi(t,\alpha))-f(t)}{\epsilon}
\end{gathered}
\end{equation}
%
%In the Euclidean space this corresponds to a one-dimensional derivative along a specified direction.  
If $f(t)$ and $g(t)$ are Linear Extended G\^ateaux differentiable functions and $\lambda$ a constant, this derivative obeys the rules:

\begin{enumerate}
	\item $d(af + bg)^{\text{leg}}(t;\psi) = ad(f)^{\text{leg}}(t;\psi) + bd(g)^{\text{leg}}(t;\psi)$
	\item $d(\lambda)^{\text{leg}}(t;\psi)=0$
	\item $d(fg)=gd(f)^{\text{leg}}(t;\psi) + fd(g)^{\text{leg}}(t;\psi)$
	\item $d\left(\frac{f}{g}\right)^{\text{leg}}(t;\psi) = \frac{gd(f)^{\text{leg}}(t;\psi) - fd(g)^{\text{leg}}(t;\psi)}{g^2}$
\end{enumerate}

%{\bf If X and Y are Euclidian space $\mathbb{R}^n$ and $\mathbb{R}^m$, respectively, $f:X\rightarrow Y$ and Linear Extended G\^ateaux differentiable at $\mathbf{x}=(x_1,x_2,\dots,x_n)\in U \subset X$, $f=(f_1,f_2,\dots,f_m)\in Y$, $\psi(x,\alpha):X\times\mathbb{R}\rightarrow X$ with $\alpha \in \mathbb{R}$}

Along this work it will be considered only functions in Euclidian space. Let $f:\mathbb{R}^n\rightarrow \mathbb{R}^m$ which is Linear Extended G\^ateaux differentiable at $\mathbf{t}=(t_1,t_2,\dots,t_n)\in T \subset \mathbb{R}^n$, {\it i. e.} it can be represented by a vector $\mathbf{f}=(f_1,f_2,\dots,f_m)\in \mathbb{R}^m$. Let $\psi(t,\alpha):\mathbb{R}^n\times\mathbb{R}\rightarrow \mathbb{R}^n$ with $\alpha \in \mathbb{R}$. Thus, the LEGD is given by\cite{zhao}

\begin{equation}\label{eq:legd22}
d\mathbf{f}^{\text{leg}}(t;\psi) = \left( \braket{\nabla f_1,\psi(t,\alpha)},\braket{\nabla f_2,\psi(t,\alpha)},\dots,\braket{\nabla f_m,\psi(t,\alpha)} \right)
\end{equation}
in which $\nabla f_j = \left(\frac{\partial }{\partial t_1}, \frac{\partial }{\partial t_2},\dots,\frac{\partial }{\partial t_n}\right)f_j$. 
%It will be considered $n=1$, for this case
Further will be considered functions $f$ such that $f_j = f_j(t)$ for $t\in\mathbb{R}$, thus $n=1$. For this particular case one obtains from equation (\ref{eq:legd22})

\begin{equation}\label{prop:5}
d(\mathbf{f})^{\text{leg}}(x;\psi) = \psi\frac{d\mathbf{f}}{dt}
\end{equation}

If $\psi$ is defined as $t^{1-\alpha}$ with $\alpha \le 1$ one obtains the conformable fractional derivative as proposed by Khalil {\it et. al.} \cite{khalil}. 
%
%\subsection{Khalil derivative}
%New definitions for fractional derivative and integral were proposed by Khalil {\it el al}\cite{khalil} 
Given a function $f:\mathbb{R}_{\ge0} \rightarrow \mathbb{R}$, its conformable fractional derivative of order $\alpha\in\left(0\right.,\left.1\right]$ is given by

\begin{equation}\label{eq:Khalil}
T_\alpha (f(t)) = \lim\limits_{\epsilon\rightarrow0} \frac{f\left( t + \epsilon t^{1-\alpha} \right) - f(t)}{\epsilon}
\end{equation}
%
%and
%\footnote{Precisa dessa equacao?}
%\begin{equation}
%T_\alpha (f(0)) = \lim\limits_{t\rightarrow0^+} T_\alpha f(t)
%\end{equation}

%{\bf property 6 applied to composite functions will be important to develop the theory for conformable neural network. For this case one can write}
Application of equation (\ref{prop:5}) is usefull in G\^ateaux-Hopfield Neural Network. Considering two functions $f:\mathbb{R}^p\rightarrow\mathbb{R}^m$ and $g:\mathbb{R}^n\rightarrow\mathbb{R}^p$, such that $f \circ g : \mathbb{R}^n\rightarrow\mathbb{R}^m$. If $n=1$ the chain rule is given by

\begin{equation}\label{eq:chain}
\begin{gathered}
d(\mathbf{f}(\mathbf{g}(t)))^{\text{leg}}(t;\psi) = \psi\left( \frac{\partial f_1}{\partial g_1}\frac{\partial g_1}{\partial t} + \frac{\partial f_1}{\partial g_2}\frac{\partial g_2}{\partial t} + \dots + \frac{\partial f_1}{\partial g_p}\frac{\partial g_p}{\partial t},\dots, \frac{\partial f_m}{\partial g_1}\frac{\partial g_1}{\partial t} + \frac{\partial f_m}{\partial g_2}\frac{\partial g_2}{\partial t} + \dots + \frac{\partial f_m}{\partial g_p}\frac{\partial g_p}{\partial t} \right) \\[3mm]
%\psi\frac{df(g(t))}{dt} = \psi\left(\frac{df}{dg}\frac{dg}{dt}\right) = \frac{df}{dg}d(g(t))^{\text{leg}}(x;\psi)
\end{gathered}
\end{equation}

%showing the chain rule for this derivative.
%Therefore, the chain rule is not an intuitive result derived from usual calculus. 
This result can be further generalized and this property will be used further in this work.

\section{G\^ateaux-Hopfield Neural Network}
The G\^ateaux-Hopfield equations will be obtained in this section. The present work will consider only linear problems, {\it i.e.} that can be represented as a transformation

\begin{equation}
\mathbf{Kf}=\mathbf{g}
\end{equation}

However, the extension to non-linear problems can be performed, as in reference \cite{jpnl}. The linear Hopfield Neural Network method consists to define a cost function as the norm of the diference between the calculated and exact results taking the derivative with respect to the learning time \cite{jpinv} as,

\begin{equation}\label{eq:lp}
\Phi(t)=\frac{1}{2}|| \mathbf{Kf} - \mathbf{g} ||^2 
\end{equation}
with $\mathbf{f}=\mathbf{f}(t)$. It is important to observe that $\Phi(t)$ can be represented as a composite function. Let $f:\mathbb{R}\rightarrow\mathbb{R}^p$ and $\phi:\mathbb{R}^p\rightarrow\mathbb{R}$, then $\Phi$ is the composite function of $\phi$ and $f$ such that $\phi\circ f:\mathbb{R}\rightarrow\mathbb{R}$. 
%Therefore if it is used the Khalil {\it et al} definition of fractional derivative with $\alpha\in\left(0,\right.\left.1\right]$, one obtains for a problem with two neurons
Under the
%Therefore if it is used the 
Linear Extended G\^ateaux Derivative definition one obtains for a problem with two neurons

\begin{equation}
d(\Phi(t))^{leg}(t;\psi) = \left(\frac{\partial \Phi}{\partial f_1}\right)_{f_2}d(f_1(t))^{leg}(t;\psi) + \left(\frac{\partial \Phi}{\partial f_2}\right)_{f_1}d(f_2(t))^{leg}(t;\psi)
\end{equation}
in which equation (\ref{eq:chain}) was applied, with $n=1$, $p=2$ and $m=1$.
%
%Since
%
%\begin{equation}
%T_\alpha(\Phi(t)) = t^{1-\alpha}\frac{d\Phi(t)}{dt}
%\end{equation}
%
%with $t\ge0$, the norm will always decrease if $\frac{d\Phi(t)}{dt}<0$. Then, 
%
For a decreasing error with time one must impose $d(\Phi(t))^{leg}(t;\psi)<0$. To satisfy this condition it is necessary to define

\begin{equation}
\begin{gathered}
d(f_1(t))^{leg}(t;\psi) = - \left(\frac{\partial \Phi}{\partial f_1}\right)_{f_2} \\[3mm]
d(f_2(t))^{leg}(t;\psi) = - \left(\frac{\partial \Phi}{\partial f_2}\right)_{f_1}
\end{gathered}
\end{equation}

Generaly the activation function, $f$, is defined as a monotonically increasing function of the neuron state, $u(t)$. Therefore one has $f_1=f(u_1(t))$ and

\begin{equation}
\begin{gathered}
d(\Phi(t))^{leg}(t;\psi) = \psi\left[\frac{d\Phi}{df_1}\frac{df_1}{du_1}\frac{du_1}{dt} + \frac{d\Phi}{df_2}\frac{df_2}{du_2}\frac{du_2}{dt}\right] = \\[3mm]
\frac{d\Phi}{df_1}\frac{df_1}{du_1}d(u_1(t))^{leg}(t;\psi) + \frac{d\Phi}{df_2}\frac{df_2}{du_2}d(u_2(t))^{leg}(t;\psi)
\end{gathered}
\end{equation}

Since $\frac{df}{du}>0$, one is left with the condition

\begin{equation}\label{eq:cond}
\begin{gathered}
d(u_1(t))^{leg}(t;\psi) = - \left(\frac{\partial \Phi}{\partial f_1}\right)_{f_2} \\[3mm]
d(u_2(t))^{leg}(t;\psi) = - \left(\frac{\partial \Phi}{\partial f_2}\right)_{f_1}
\end{gathered}
\end{equation}

%Rewriting the left hand side as $T_\alpha(u_i(t))=t^{1-\alpha}\frac{du}{dt}$.  %and noting that $t>0$ one must impose
%
%\begin{equation}\label{eq:cond2}
%\begin{gathered}
%\frac{du_1(t)}{dt} = - \left(\frac{\partial \Phi}{\partial f_1}\right)_{f_2} \\[3mm]
%\frac{du_2(t)}{dt} = - \left(\frac{\partial \Phi}{\partial f_2}\right)_{f_1}
%\end{gathered}
%\end{equation}
%
Using equation (\ref{eq:lp}), one obtains

\begin{equation}
\Phi(t) = \frac{1}{2}\left[ \left(K_{11}f_1 + K_{12}f_2 - g_1\right)^2 + \left(K_{21}f_1 + K_{22}f_2 - g_2\right)^2 \right]
\end{equation}
and therefore

%\begin{equation}
%\begin{gathered}
%\frac{du_1(t)}{dt} = - K_{11}\left( K_{11}f_1 + K_{12}f_2 - g_1\right) - K_{21}\left( K_{21}f_1 + K_{22}f_2 - g_2 \right) \\[3mm]
%\frac{du_2(t)}{dt} = - K_{12}\left( K_{11}f_1 + K_{12}f_2 - g_1\right) - K_{22}\left( K_{21}f_1 + K_{22}f_2 - g_2 \right)
%\end{gathered}
%\end{equation}
%
%or, more generally

%\begin{equation}\label{eq:hop}
%\frac{d\mathbf{u}(t)}{dt} = -\mathbf{K^TKf} + \mathbf{K^Tg}
%\end{equation}
%
%Then one has

\begin{equation}\label{eq:2x2}
\begin{gathered}
d(u_1(t))^{leg}(t;\psi) = - K_{11}\left( K_{11}f_1 + K_{12}f_2 - g_1\right) - K_{21}\left( K_{21}f_1 + K_{22}f_2 - g_2 \right) \\[3mm]
d(u_2(t))^{leg}(t;\psi) = - K_{12}\left( K_{11}f_1 + K_{12}f_2 - g_1\right) - K_{22}\left( K_{21}f_1 + K_{22}f_2 - g_2 \right)
\end{gathered}
\end{equation}

The set of equations (\ref{eq:2x2}) can be rearranged and written in a matricial form, which is the same for any dimension,

\begin{equation}\label{eq:hopdfrac}
d(\mathbf{u}(t))^{leg}(t;\psi) = -\mathbf{K^TKf} + \mathbf{K^Tg}
\end{equation}
or

\begin{equation}\label{eq:GHNN}
\frac{d\mathbf{u}}{dt} = \frac{1}{\psi}\left[ -\mathbf{K^TKf} + \mathbf{K^Tg} \right]
\end{equation}
%
%in which setting 
For $\psi=1$ the classical set of differential equations for Hopfield Neural Network is recovered. In this work it will be chosen $\psi = t^{1-\alpha}$ as in Khalil {\it et al} definition, but here $\alpha$ is a parameter and can be any real number. \\
%Since $t>0$ in $T_\alpha(u)=t^{1-\alpha}\frac{du}{dt}$, one may define $\frac{du}{dt} = -\frac{\partial \Phi}{\partial f}$ in equation (\ref{eq:cond}) and obtains the usual Hopfield Neural Network method instead. 
%It will be analysed how this new definition affects the method. 

Considering equation (\ref{eq:GHNN}) and this particular for $\psi$ it is possible to propagate the fractional Hopfield equations by solving

\begin{comment}
\section{Conformable Hopfield Neural Network equations: the fractional order case}
An alternative option is not to drop off the term $t^{1-\alpha}$ passing from equation (\ref{eq:cond}) to equation (\ref{eq:cond2}). Then one has

\begin{equation}
\begin{gathered}
T_\alpha(u_1(t)) = - K_{11}\left( K_{11}f_1 + K_{12}f_2 - g_1\right) - K_{21}\left( K_{21}f_1 + K_{22}f_2 - g_2 \right) \\[3mm]
T_\alpha(u_2(t)) = - K_{12}\left( K_{11}f_1 + K_{12}f_2 - g_1\right) - K_{22}\left( K_{21}f_1 + K_{22}f_2 - g_2 \right)
\end{gathered}
\end{equation}

Rewriting in a matricial form

\begin{equation}\label{eq:hopdfrac}
T_\alpha(\mathbf{u}(t)) = -\mathbf{K^TKf} + \mathbf{K^Tg}
\end{equation}
%
which is validy for any case. Setting $\alpha=1$ it is obtained the classical set of differential equations for Hopfield Neural Network. In this work it will be analysed how this new definition affects the method. 
Considering equation (\ref{eq:hopdfrac}) it is possible to propagate the fractional Hopfield equations by solving
\end{comment}

\begin{equation}\label{eq:hopfrac}
\frac{d\mathbf{u}}{dt} = \frac{1}{t^{1-\alpha}}\left[ -\mathbf{K^TKf} + \mathbf{K^Tg} \right]
\end{equation}

%This equation is equivalent to the ordinary HNN method with a time scale, it will be clarified further in this work. 
To avoid singularity, the initial condition must be given at $t_0>0$.
%Since the initial guess does not has to begin at $t=0$ one may solve this set of differential equations by choosing any initial time $t_0>0$.

%In this work the equations were solved by the fourth order Runge-Kutta with variable step with $f(u)=110\tanh(u)$ as activation function in all cases.
%{\bf Nao precisa desses detalhes, {\tt ode45} function, in MATLAB\textsuperscript{\textregistered} enviroment 
	%with $f(u)=110\tanh(u)$ as activation function in all cases.
%	}
%\clearpage
%\section{Example on first order Fredholm integral equation}
\section{The prototype system}
%\subsection{Apresentado o prototipo}
Application and testing of the conformable Hopfield neural network will be carried out with a Fredholm integral equation of first order,

\begin{equation}
g(x) = \int_{a}^{b}K(x,y)f(y)dy
\end{equation}
with
	
\begin{equation}
\begin{gathered}
K(x,y)=(x+y)^{-1} \\[3mm]
f(y)=y^{-1} \\[3mm]
g(x)=x^{-1}\ln\left(\frac{1+\frac{x}{a}}{1+\frac{x}{b}}\right)
\end{gathered}
\end{equation}
in which $a=1$ and $b=5$. This prototype integral equation was used before to discuss the singular value decomposition and Tikhonov regularization \cite{riele,jp1}.
	
%To ilustrate an application it will be solved the same first order Fredholm integral equation presented by Riele \cite{riele}. In this example one has
%\subsection{Detalhes da base}

The problem was well described by $R^n$ and $R^m$ with $m=n=22$. The condition number, $\kappa(\mathbf{K})$, that indicates how a small variation in $\mathbf{g}$ can amplify the error in $\mathbf{f}$ and is given by

\begin{equation}
\kappa(\mathbf{K})=\frac{\sigma_{max}(\mathbf{K})}{\sigma_{min}(\mathbf{K})}
\end{equation}
with $\sigma_{{max}}(\mathbf{K})$ and $\sigma_{{min}}(\mathbf{K})$ the maximum and minimum singular values of $\mathbf{K}$, respectively. For this case the condition number is equals to $8.1167\times10^{18}$, which defines an ill-posed problem. 
%It will be analysed the differences between solving this problem with usual ($\alpha=1$) and fractional ($\alpha<1$) Hopfield equations. The initial condition for the neuron states was defined such that one has $f(u_i)=0.4$ for all i.
%\clearpage
\section{The numerical solution}
%The new HNN method for different values of fractional derivative order ($\alpha \in \left(0,\right.\left.1\right]$) were applied to the Fredholm integral of first kind problem presented previously, but for 8 neurons in which the initial condition for the neuron states were the same. The results are presented in Figure \ref{fig:Rie1}

%\begin{figure}[h!]\centering
%	\includegraphics[width=8cm]{normsRie.eps}
%	\caption{Norm calculated for $\alpha$ from $0.1$ (upper line) to $0.9$ (lower line).}\label{fig:Rie1}
%\end{figure}
%\subsection{Converg\^encia para ponto fixo e LEGD-Euler method}
Considering equation (\ref{eq:hopfrac}) the fixed point for this set of differential equations, {\it i.e.} $\frac{d\mathbf{u}}{dt} = 0$, is $\mathbf{u}$ such that

\begin{equation}
\mathbf{K}\mathbf{f}(\mathbf{u}(t)) = \mathbf{g}
\end{equation}

This result is independent of $\alpha$, but this parameter has influence on the convergence to fixed points. 
% as seen in the previous section. 
Since the conformable fractional derivative is a particular case of LEGD one may maintain $\psi(t,\alpha)=t^{1-\alpha}$, but without the restriction of $\alpha<1$.
%in examples presented in Apendix \ref{ap:1}. 
%Therefore one may think of setting 
%Therefore, now it is possible to set $\alpha>1$.
% to decreases the learning time. 
To help understanding the influence of $\alpha$ on the convergence one may consider the numerical solution of the GHNN by the Euler method, 

\begin{equation}
\mathbf{u}_{i+1} = \mathbf{u}_i + \frac{h}{t_i^{1-\alpha}}\left[ \mathbf{K'Kf}_i - \mathbf{K'g} \right]
\end{equation}

Defining $\frac{h}{t^{1-\alpha}}=\epsilon$ one has

\begin{equation}\label{eq:fracEul}
\mathbf{u}_{i+1} = \mathbf{u}_i + \epsilon\left[ \mathbf{K'Kf}_i - \mathbf{K'g} \right] = \mathbf{u}_i + \epsilon\left(\frac{d\mathbf{u}}{dt}\right)_{\text{usual HNN}}
\end{equation}
which is analogous to the usual Euler method with a time dependent step. For $\alpha < 1$ one has $\epsilon$ decrescent with time and therefore the next point is increasingly closer to the previous point, which explains the need to define a larger time of integration. For $\alpha > 1$ one has $\epsilon$ crescent and hence the next point obtained in the integration is farther from the previous point, which explains the lower learning time. More generally, one may write

\begin{equation}\label{eq:legd:eul}
\mathbf{u}_{i+1} = \mathbf{u}_i + \frac{h}{\psi(t_i;\alpha)}\left[ \mathbf{K'Kf}_i - \mathbf{K'g} \right]
\end{equation}
which may be called LEGD-Euler method.\\

%\subsection{%Justificativa e comparacao com passo variavel
%Interpreta\c{c}\~ao como modifica\c{c}\~ao da taxa de aprendizagem cl\'assica}

Since the time-step in the LEGD-Euler method is constant and defined by $h$, the analogy with a time dependent step method must be carefully interpreted. As Zhao and Luo stated in their work\cite{zhao}, the LEGD is a modification of the classical velocity both in direction (for higher dimensions) and magnitude. From this point of view, the conformable Hopfield set of differential equations can be interpreted as a modification of the classical learning rate magnitude. \\

\section{
	%Application of new HNN with $\alpha>1$ to the first order Fredholm integral
	%Order theoretical analysis on the Hopfield solution 
	Convergence dependence on $\alpha$ and $\psi$
}\label{sec:timedep}

From this arguments, the learning rate dependence on $\alpha$ and $\psi$ will be possible to be predicted. For $\psi=t^{1-\alpha}$ and $\alpha<1$ the classical learning rate will decrease over time and the neural network will have a fast learning only in the beginning of the process. If the initial condition is not near to the exact solution, the neural network will need a larger time to learn about the process and for even larger time it will can not learn anymore since $\psi=0$. For $\alpha>1$ the classical learning time will be modified to zero on the beginning, however after a short time it will monotonically increase with time until the neural network has learnt completely about the process, which will demand a lesser time if compared with the usual HNN method.\\

%\section{CPU time for $\alpha \ge 1$}
%\clearpage
%\section{Convergence analysis}
Since the conformable Hopfield differential equations can be interpreted as a 
modification 
of the classical Hopfield equations, it will change the learning and computational time for  the calculation. The effect of $\alpha  \ge 1$ on the computational time will be carried out.\\

%To analyse the effect of $\alpha$ on this computational time it will be used the LEGD-Euler method to solve the first order Fredholm integral inverse problem with 22 neurons, with $\alpha \ge 1$.\\ %, since it was observed a smaler learning time in this case. 
%\clearpage
%\subsection{Tabela 1 e figura das normas}
%\section{Norm convergence and derivative order}
\section{Learning time evolution}
\begin{figure}[h!]\centering
	\includegraphics[width=8cm]{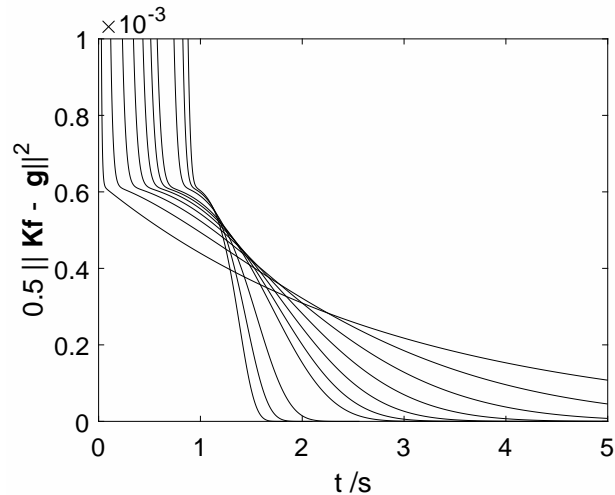}
	\caption{Norm for $\alpha$ equals to $1$, $1.5$, $2$, $2.5$, $3$, $3.5$, $4$, $6$, $8$ and $10$.}\label{fig:10}
\end{figure}

The results obtained, for a fixed norm of about $10^{-11}$, are presented in Table \ref{tab:3} whereas
in 
Figure {\ref{fig:10}}  the norm times evolution is presented. From these results one can observe a smaller computational time to obtain the norm of the same magnitude as $\alpha$ increases. 
\\
\begin{table}[h!]\centering\caption{Computational time dependency on $\alpha$.
		%to obtain $10^{-11}$.
	}\label{tab:3}
	\begin{tabular}{cc}
		\hline
		$\alpha$ & Time /s \\
		\hline
		1   & 5.828122 \\
		1.5 & 2.090320 \\
		2   & 1.244758 \\
		2.5 & 0.887580 \\
		3   & 0.706472 \\
		3.5 & 0.659272 \\
		4   & 0.509641 \\
		6  & 0.356831 \\
		8  & 0.298397 \\
		10 & 0.263862 \\
		\hline
	\end{tabular}
\end{table}
\FloatBarrier

\begin{figure}[h!]\centering
	\includegraphics[width=8cm]{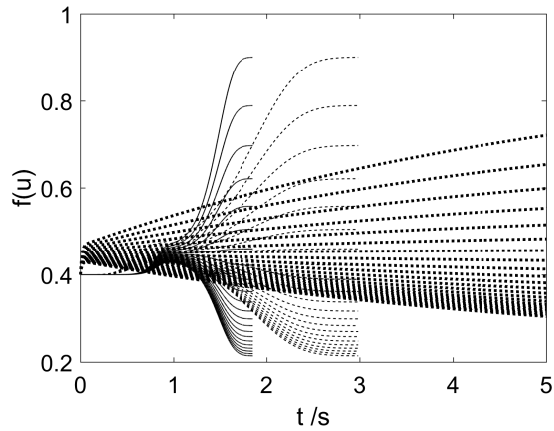}
	\caption{Solution evolution in time for $\alpha$ equals to $1$ (dotted), $5$ (dashed) and $10$ (full line).}\label{fig:11}
\end{figure}

\FloatBarrier
%\clearpage
%
%it is given the time evolution of the norm, 
%for the interval $t\in[0,5]$.
%, for each case. 
%The maximum norm value 
%%at y-axis 
%was set to $10^{-3}$ to a better visualization of the behaviour.
%\clearpage
%\subsection{Evolucao dos neurônios}
%\section{Learning time evolution}
To further 
validate
%enriched
%ilustrate 
%corroborate
the discussion 
%presented 
%in section \ref{sec:timedep} 
it is presented in Figure \ref{fig:11} the solution as a function of learning time evolution for $\alpha=1$, $5$ and $10$. 
As predicted, for $\alpha>1$ the network will not learn in the begining of the process. However after the learning process starts, the convergency to the exact solution is faster for larger values of $\alpha$. 
The final results were the same in all calculations, but with the advantage of getting the results about 22 times faster for $\alpha=10$. From Figure \ref{fig:10} it is also possible to see a limit for learning time as $\alpha$ increases. \\
%time will not decresce indefinitely with the increase of $\alpha$.

\section{Conclusions}
A modified Hopfield neural network was discussed in the present work from the conformable derivative point of view. Properties of this derivative were enunciated and applied to establish a new set of Hopfield differential equations.
% 
%In this work it were presented the definitions of LEGD and conformable fractional %derivative, which is a particular case of the former. Some properties were listed, the %most of them keeping the usual calculus form. \\
%An example presented by Khalil {\it et. al.} was solved numerically and compared with the analytical solution. \\
%Since the chain rule in conformable fractional derivative has a similarity to the usual calculus this propertie was used to obtain the conformable HNN, for which the usual method is retrieved by setting $\alpha=1$. 
The method was implemented to solve a first order Fredholm integral problem and the results were compared with the usual Hopfield Neural Network. \\
%It was observed the needing of a larger time integration for $\alpha<1$. 
%The influence of $\alpha$ on the convergence to the fix point by the definition of the LEGD-Euler method were also discussed.\\

%It was analysed how $\alpha$ influnces on the convergency to the fix point by the definition of the LEGD-Euler method. 

The numerical solution 
%of this conformable Hopfield neural network 
can be related to the usual Euler method with a time-dependent step, though this must be done carefully, since time step, $h$, used in the numerical procedure is constant. Another interpretation can be obtained considering the LEGD as a modification of the classical learning rate.\\
%classical velocity  
%both in direction and magnitude. \\

%Thus, the conformable HNN method is the modification of the classical Hopfield set of differential equations. 

%This behaviour is in agreement with the prediction made by interpreting the new method as a change of the classical learning rate. 
The advantage and robustness of this new method is evident since for the present problem it was found better results for larger values of $\alpha$, which gives the exact solution up to 22 times faster than the usual HNN method. 
%Further studies considering different activation energies, $\psi(\alpha,t)$ and initial conditions will be performed to further explore this method. 

\section*{Acknowledgments} 
We would like to thank CNPq for the financial support.

\end{document}